\documentclass[journal]{IEEEtran}

\usepackage[T1]{fontenc}
\usepackage[utf8]{inputenc}
\usepackage{amsmath,amssymb}
\usepackage{graphicx}
\usepackage{booktabs}
\usepackage{tabularx}
\usepackage{array}
\usepackage{multirow}
\usepackage{cite}
\usepackage{microtype}
\usepackage{url}
\usepackage{xcolor}
\usepackage{enumitem}

\graphicspath{{figures/}}

\newcolumntype{Y}{>{\raggedright\arraybackslash}X}

\newcommand{\figwidth}{0.99\textwidth}

\newcommand{\HeshamEmail}{hrakha@vt.edu}
\newcommand{\MonaEmail}{m.jaber@qmul.ac.uk}
\newcommand{\MoussaEmail}{mayyash@csu.edu}

\author{Lyes~Saad~Saoud,
        Hesham~A.~Rakha, \emph{Fellow, IEEE} ,
        Mona~Jaber,
        and~Moussa~Ayyash%
\thanks{Lyes Saad Saoud and Moussa Ayyash are with the Center for Information \& Security Education and Research, Chicago State University, Chicago, IL 60628 USA (e-mail:  \MoussaEmail).}%
\thanks{Hesham A. Rakha is with the Center for Sustainable Mobility, Virginia Tech Transportation Institute, Blacksburg, VA 24061 USA (e-mail: \HeshamEmail).}%
\thanks{Mona Jaber is with the School of Electronic Engineering and Computer Science, Queen Mary University of London, London E1 4NS, U.K. (e-mail: \MonaEmail).}%
\thanks{Corresponding author: Lyes Saad Saoud.}%
}

\title{Beyond Dead Reckoning: A Point of View on Camera--DAS--GNSS Continuity in Road Tunnels}

\begin{document}
\bstctlcite{IEEEexample:BSTcontrol}
\maketitle

\begin{abstract}
Road tunnels remove satellite visibility where connected and automated vehicles still require continuous, attributable, and integrity-bounded positioning. This Point of View argues that tunnel localization should be treated as infrastructure-assisted cross-modal track continuity, not as extrapolation from the last trusted satellite fix. A trusted portal satellite solution provides the global anchor, distributed acoustic sensing (DAS) continuous motion evidence, cameras sparse identity and lane anchors, onboard sensing short-term dynamics, and edge computing association, fusion, integrity monitoring, and guarded reacquisition. Ground truth is reserved for offline calibration and validation. The article develops a falsifiable research and deployment agenda for progressing from synchronized multimodal evidence to validated, integrity-aware tunnel positioning continuity.
\end{abstract}


\section{The Missing Fix Is Not the Whole Problem}
\IEEEPARstart{R}{oad} tunnels expose a structural weakness in vehicle localization. Satellite positioning can deteriorate near a portal and disappear inside the tunnel, while automated and connected functions still require a continuously referenced state. Inertial sensing, wheel odometry, visual--inertial estimation, lidar localization, and map constraints can bridge part of the gap, but bias, slip, repetitive geometry, illumination, and multipath each create failure modes \cite{campos2021orbslam3,jiang2025tunnel,kuutti2018survey,qin2018vins}.

The usual question is therefore too narrow: \emph{How accurately can the next positions be extrapolated after the final trusted satellite fix?} A smooth trajectory can still belong to the wrong vehicle or lane, map to the wrong fiber branch, arrive too late for control, or be accepted with unjustified confidence. Identity continuity, geometry, timing, uncertainty, and integrity are inseparable.

This article advances the following position:
\begin{quote}
\emph{Tunnel positioning should be formulated as a cross-modal track-continuity and integrity service. A trusted portal satellite solution supplies the global anchor; distributed acoustic sensing supplies continuous along-corridor physical evidence; calibrated cameras supply sparse identity and lane anchors; onboard motion sensing supplies short-term dynamics; and an edge estimator associates these complementary observations to preserve each vehicle track while exposing its state, uncertainty, and service status.}
\end{quote}

The novelty of this position is therefore not the coexistence of multiple sensing modalities, but the requirement that global reference, vehicle identity, uncertainty, integrity, and guarded reacquisition be treated and validated jointly as properties of a single positioning-continuity service. If the entry satellite solution cannot satisfy the required integrity checks, the infrastructure may still support relative track continuity, but it cannot claim globally referenced continuity from that anchor.

\begin{figure*}[!t]
  \centering
  \includegraphics[width=\figwidth]{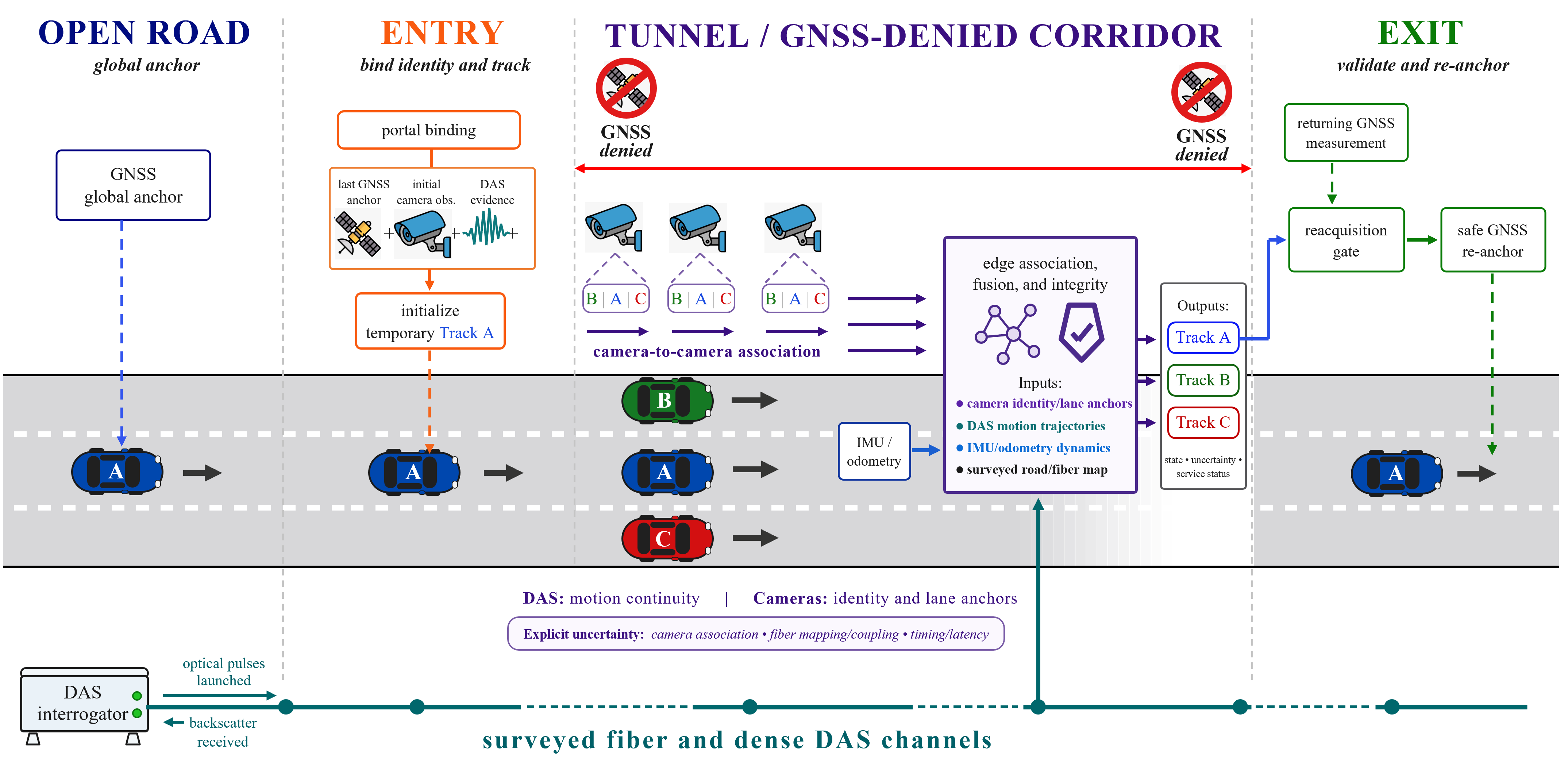}
  \caption{Reference architecture for infrastructure-assisted track continuity in a GNSS-denied road tunnel. A trusted portal satellite solution anchors the track; cameras provide identity and lane observations; DAS provides distributed motion evidence; onboard motion and surveyed road--fiber geometry constrain the state; and the edge estimator outputs tracks with uncertainty and service status. Returning satellite measurements are admitted only after a consistency-based reacquisition gate. The figure is conceptual and reports no experimental results.}
  \label{fig:architecture}
\end{figure*}

Figure~\ref{fig:architecture} summarizes the information flow. Entry observations initialize a temporary track; inside, cameras preserve identity while DAS, onboard motion, and surveyed geometry constrain the trajectory; at exit, returning satellite measurements pass a consistency gate before re-anchoring.

The architecture is a technology position, not a performance claim. Five distinctions are central: the entry satellite anchor must itself satisfy the required trust checks; fiber coordinates require surveyed road registration; camera-to-camera and camera--DAS association solve different identity problems; all sources carry uncertainty; and a returning satellite fix must be checked before trust is restored.

\section{Article Boundary and Evidence Base}
This Point of View asks what each modality observes, how observations share a road frame and identity, how uncertainty and integrity are exposed, and what evidence supports progressively stronger claims. It synthesizes evidence from tunnel localization \cite{jiang2025tunnel,kuutti2018survey}, fiber traffic sensing \cite{chiang2023das,hartog2017das,yuan2024spatial,zhan2020das}, infrastructure vision \cite{cui2024vilam,tang2019cityflow,zhang2000calibration}, and fusion, integrity, and safety \cite{barshalom2001estimation,brown1992raim,dellaert2017factor,guo2017calibration,iso21448,iso26262}. Camera-supervised fiber monitoring and traffic-assisted fiber geolocation provide important calibration evidence \cite{cohen2025video,cohen2026fibergeo}, but not yet an operational tunnel service.

\section{What the Modalities Can and Cannot Observe}
\subsection{Satellite positioning and onboard dead reckoning}
A trusted satellite solution at the portal provides the global coordinates and time used to anchor the service \cite{kaplan2017gnss}. Onboard inertial, wheel, visual, lidar, and vehicle-dynamic sources bridge outages but drift; infrastructure should therefore complement, not silently overwrite, the vehicle estimate.

\subsection{Distributed acoustic sensing as a continuous backbone}
A DAS interrogator turns fiber into dense vibration-sensitive channels \cite{hartog2017das,zhan2020das}; vehicle trajectories in time--distance space support detection, speed, classification, and reconstruction \cite{chiang2023das,yuan2024spatial}. Geographic positioning requires a road--fiber map from survey, known access points, tap testing, or traffic-assisted calibration \cite{cohen2026fibergeo}. Its uncertainty must be versioned and revalidated. DAS is a longitudinal backbone, not a complete identity or lane sensor.

\subsection{Cameras as sparse semantic anchors}
Cameras provide appearance, lane, direction, class, and calibrated road coordinates \cite{cui2024vilam,zhang2000calibration}; multi-camera tracking can preserve identity \cite{bewley2016sort,tang2019cityflow,wojke2017deepsort}. Their role is to label and correct anonymous fiber tracks. Occlusion, lighting, compression, and camera movement remain explicit failures; plate or face recognition is unnecessary.

\begin{table*}[!t]
\caption{Complementary Observability and Failure Modes}
\label{tab:modalities}
\centering
\footnotesize
\begin{tabularx}{\textwidth}{@{}p{0.12\textwidth}p{0.22\textwidth}p{0.26\textwidth}Y@{}}
\toprule
\textbf{Source} & \textbf{Primary observable} & \textbf{Dominant limitation} & \textbf{Role in the position-continuity service} \\
\midrule
Satellite positioning & Global position, velocity, time & Obstruction, multipath, interference, spoofing, portal transition & Candidate global anchor before entry and re-anchor after exit, subject to integrity checks \\
Onboard inertial/odometry & High-rate relative motion & Bias, slip, model error, drift & Motion factor and independent vehicle-side consistency check \\
Fixed cameras & Identity-bearing tracklet, lane, direction, calibrated road coordinate & Occlusion, illumination, limited coverage, camera displacement & Sparse semantic and geometric anchor; track initialization and correction \\
Distributed acoustic sensing & Continuous along-fiber disturbance track, speed and direction evidence & Anonymous tracks, coupling changes, overlap, fiber-to-road ambiguity & Continuous infrastructure motion backbone between camera anchors \\
Surveyed map and fiber geometry & Road centerline, lanes, portals, cable route and orientation & Static information; survey and maintenance errors & Common coordinate frame, feasibility constraints, channel-to-road mapping \\
Vehicle-to-infrastructure link & Timestamped state and service information & Latency, packet loss, authentication, age of information & Delivery of state, uncertainty, integrity mode, source health, validity horizon \\
\bottomrule
\end{tabularx}
\end{table*}

Table~\ref{tab:modalities} emphasizes that the architecture is not a competition among sensors. Its value comes from complementary observability and from declaring degradation when that complementarity disappears.

\section{A Reference Architecture for Position Continuity}
\subsection{State and surveyed geometry}
For a road-constrained tunnel, a useful state is
\begin{equation}
\begin{aligned}
  \mathbf{x}_t &= [s_t,\, \ell_t,\, v_t,\, a_t,\, \psi_t]^{\mathsf T},\\
  \mathbf{p}_t &= \mathcal{R}(s_t,\ell_t),
  & s_t &= \mathcal{M}_{\mathrm f}(q_t).
\end{aligned}
\label{eq:state}
\end{equation}
Here $s_t$ is road chainage, $\ell_t$ lateral offset, $v_t$ and $a_t$ speed and acceleration, $\psi_t$ heading, and $q_t$ a fiber coordinate. $\mathcal{M}_{\mathrm f}$ maps fiber to road chainage and $\mathcal{R}$ to the selected frame. Uncertainty in $\mathcal{M}_{\mathrm f}$ must be propagated into the state uncertainty or integrity bound rather than treated as fixed geometry. Sensor overlap at entry and exit binds the vehicle to camera/acoustic tracks and predicts where a returning satellite solution should appear.

\subsection{A service output, not only a coordinate}
A tunnel edge node should output at least
\begin{equation}
  \mathcal{Y}_t = \{\hat{\mathbf{x}}_t,\,\mathbf{P}_t,\,\mathcal{I}_t,\,
  \gamma_t,\,\mathcal{H}_t,\,t_t,\,T_{\mathrm{valid}}\},
  \label{eq:service}
\end{equation}
where $\mathbf{P}_t$ is a covariance or other defensible uncertainty representation, $\mathcal{I}_t$ a temporary identity, $\gamma_t$ association confidence, $\mathcal{H}_t$ integrity/source health, $t_t$ timestamp, and $T_{\mathrm{valid}}$ validity horizon. The frame must be explicit and the message authenticated before V2I use \cite{threegpp22186}. Such an output lets the vehicle reject stale information, compare infrastructure and onboard estimates, widen safety margins, or enter a minimum-risk mode.

\section{Cross-Modal Association Is the Core Intelligence}
The central question is: \emph{Which camera tracklet and which acoustic trajectory belong to the same vehicle?} Geometrically plausible trajectories are unsafe if identities are exchanged.

\begin{figure*}[!t]
  \centering
  \includegraphics[width=0.96\textwidth]{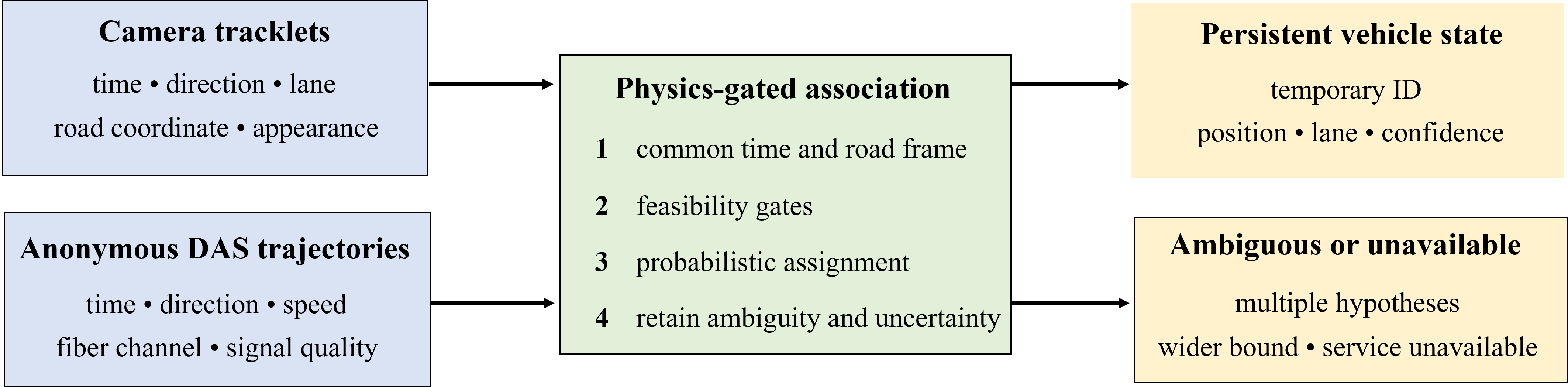}
  \caption{Cross-modal association as an identity-continuity problem. Camera tracklets and anonymous fiber tracks enter a physics-gated probabilistic matcher. A unique persistent track is emitted only when timing, direction, geometry, dynamics, and uncertainty are mutually consistent. Otherwise, the system retains multiple hypotheses, widens the bound, or declares the service unavailable. The figure is conceptual and contains no generated scores or results.}
  \label{fig:association}
\end{figure*}

For camera tracklet $i$ and acoustic track $j$, a transparent association cost is
\begin{equation}
 C_{ij} = \sum_{k \in \mathcal{F}} w_k\,
 \rho_k\!\left(\Delta z_{ij,k};\,\Sigma_{ij,k}\right),
 \quad \mathcal{F}=\{t,v,d,s\},
 \label{eq:association}
\end{equation}
where the shared cross-modal features represent time, speed, direction, and projected road position. Lane consistency should be included only where the installed fiber geometry and calibration make lane observability defensible; appearance features belong to camera-to-camera identity continuity rather than direct camera--DAS matching. Hard physical gates should precede learned scoring. When evidence is ambiguous, the system should expose multiple hypotheses or assignment uncertainty rather than force a confident identity \cite{barshalom2001estimation,brambilla2020cooperative}. Traversal-specific pseudonyms and local feature processing can reduce privacy exposure.

\section{Fusion, Integrity, and Guarded Reacquisition}
\subsection{A transparent fusion formulation}
A factor graph or Bayesian smoother keeps each constraint visible \cite{dellaert2017factor,sarkka2013bayesian}. For a trajectory $\mathbf{X}=\{\mathbf{x}_0,\ldots,\mathbf{x}_T\}$,
\begin{equation}
\begin{aligned}
\hat{\mathbf{X}} = \arg\min_{\mathbf{X}}\; &
\sum_t \|\mathbf{r}^{\mathrm{mot}}_t\|^2_{\mathbf{W}_{\mathrm m}}
+\sum_t \|\mathbf{r}^{\mathrm{das}}_t\|^2_{\mathbf{W}_{\mathrm d}}\\
&+\sum_k \|\mathbf{r}^{\mathrm{cam}}_k\|^2_{\mathbf{W}_{\mathrm c}}
+\sum_t \|\mathbf{r}^{\mathrm{map}}_t\|^2_{\mathbf{W}_{\mathrm r}}\\
&+\|\mathbf{r}^{\mathrm{portal}}\|^2_{\mathbf{W}_{\mathrm p}}.
\end{aligned}
\label{eq:fusion}
\end{equation}
Equation~\eqref{eq:fusion} is illustrative rather than prescriptive: it exposes the factorization required by the proposed service without prescribing a particular estimator, residual distribution, or independence assumption. The formulation is conditional on an association hypothesis; when association is ambiguous, the estimator should retain competing hypotheses or explicitly represent assignment uncertainty rather than silently selecting one correspondence. Cross-modal errors may be correlated, non-Gaussian, and operating-condition dependent, requiring robust factors, conservative covariance treatment, or explicit multi-hypothesis inference. The information matrices must reflect occlusion, delay, weak coupling, and domain shift. Learned residual or noise models remain useful only if their uncertainty is empirically calibrated \cite{guo2017calibration,kendall2017uncertainty}.

\begin{figure*}[!t]
  \centering
  \includegraphics[width=0.90\textwidth]{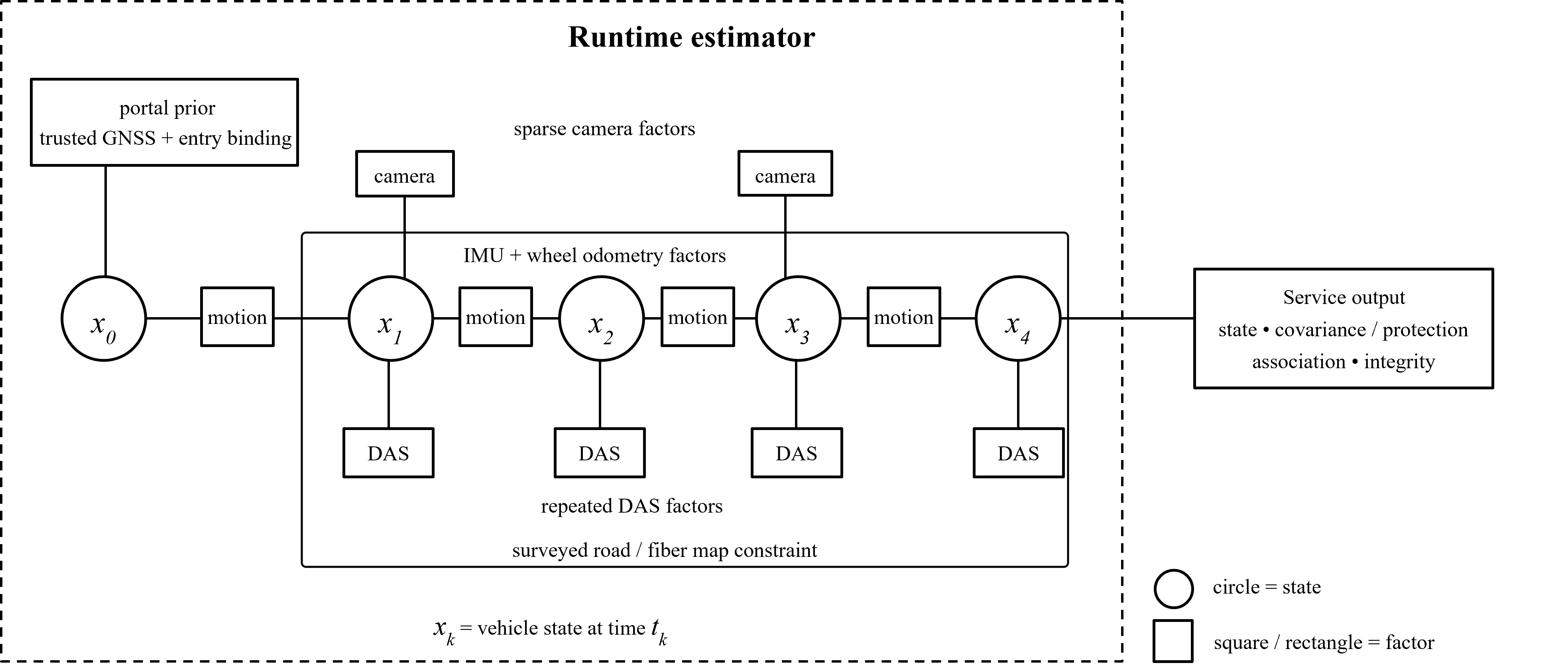}
  \caption{Transparent factorization of the tunnel trajectory. A trusted entry satellite solution, candidate exit satellite measurements, sparse camera anchors, continuous acoustic tracks, a surveyed road/fiber map, and optional onboard motion constrain a shared state sequence. The operational output contains pose, uncertainty or a candidate protection bound, association state, and integrity state. Independent ground truth is used offline to calibrate and audit these factors; it is not a runtime measurement.}
  \label{fig:factorgraph}
\end{figure*}

\subsection{Integrity is a declared operating condition}
Let $PL_t$ denote a candidate conservative error bound or, once validated, a protection level, and let $AL_t$ denote the application alert limit. A minimal rule is
\begin{equation}
\mathcal{H}_t =
\begin{cases}
\text{nominal}, & PL_t \leq AL_t,\ \text{checks pass},\\
\text{degraded}, & PL_t > AL_t\ \text{or reduced observability},\\
\text{fallback}, & \text{identity, timing, or bound invalid}.
\end{cases}
\label{eq:integrity}
\end{equation}
The exact protection computation remains an open requirement. Until empirical coverage and integrity-risk properties are established, $PL_t$ should be interpreted as a candidate conservative bound rather than an operationally validated protection level. Integrity language is not justified without empirical coverage, false-safe probability, and time-to-alert evidence, particularly for correlated, non-Gaussian camera--fiber errors \cite{brown1992raim}.

\subsection{Reacquisition is a transition, not a switch}
A returning satellite solution should first be treated as a candidate measurement source and checked against predicted position and motion, lane and map feasibility, timing, uncertainty, and recent infrastructure evidence. Let $\mathbf{z}^{\mathrm{sat}}_t$ denote the returning satellite measurement and $h_{\mathrm{sat}}(\hat{\mathbf{x}}^{\mathrm{tun}}_t)$ its prediction from the tunnel estimate. A compact measurement-space consistency statistic is
\begin{equation}
\boldsymbol{\nu}_t=\mathbf{z}^{\mathrm{sat}}_t-h_{\mathrm{sat}}(\hat{\mathbf{x}}^{\mathrm{tun}}_t),
\qquad
d_t^2=\boldsymbol{\nu}_t^{\mathsf T}\mathbf{S}_t^{-1}\boldsymbol{\nu}_t,
\label{eq:reacqgate}
\end{equation}
where $\mathbf{S}_t$ includes the satellite-measurement and tunnel-prediction uncertainty together with known cross-correlation, and is conservatively inflated or bounded when dependence is uncertain. Satellite measurements should be admitted to the estimator only after $d_k^2\leq\tau$ for $k=t-N+1,\ldots,t$ and the map, motion, timing, and source-health gates pass. Once admitted, their influence can increase gradually through estimator weighting appropriate to the measurement model and covariance rather than by directly blending state vectors. The window $N$ and threshold $\tau$ are operating-domain-dependent quantities that must be calibrated against false-reacquisition probability, time to re-anchor, and the downstream alert limit. Persistent disagreement therefore reduces trust or triggers fallback rather than forcing the tunnel trajectory to snap to an unverified fix.

\section{Ground Truth Is a Development Instrument}
Ground truth is an offline development instrument for fiber registration, association labels, noise and integrity calibration, domain-shift detection, and audit. The reference trajectory must be sufficiently independent of the evaluated cameras, fiber map, and learned features. Traffic-assisted fiber geolocation is therefore not independent truth for the complete service; when camera observations help estimate $\mathcal{M}_{\mathrm f}$, a separate reference is needed to expose common-mode error. Privileged truth must not leak into deployed inputs or evaluation splits.

\section{Failure Modes Hidden by Average Position Error}
Mean trajectory error cannot establish a safe service. Table~\ref{tab:failures} lists failures that can coexist with an apparently smooth path.

\begin{table*}[!t]
\caption{Critical Failure Hypotheses and Required Evidence}
\label{tab:failures}
\centering
\footnotesize
\begin{tabularx}{\textwidth}{@{}p{0.19\textwidth}p{0.31\textwidth}Y@{}}
\toprule
\textbf{Failure hypothesis} & \textbf{Why accuracy-only evaluation can miss it} & \textbf{Required defense and test} \\
\midrule
Track swap in dense traffic & The set of reconstructed paths remains geometrically plausible while paths are assigned to the wrong vehicles. & Identity-switch rate, association entropy, track completeness, controlled close-following and overtaking trials, oracle-association baseline. \\
Fiber-path ambiguity & A channel is mapped to the wrong road segment, lane influence, or direction, especially near loops, crossings, and changing offsets. & Surveyed fiber digital twin, direction tests, map residuals, map-version control, perfect-map counterfactual. \\
Heavy vehicle masks light vehicle & Aggregate flow remains accurate while an individual car or motorcycle disappears. & Per-class and per-lane recall, mixed truck/passenger scenarios, missed-track alert behavior. \\
Camera occlusion or lighting failure & The estimator looks continuous because it silently ignores a missing identity anchor. & Explicit source-availability state, camera blackout, glare, darkness, smoke, dirt, and compression tests. \\
Changing fiber coupling & A model remains accurate on training data but becomes miscalibrated after water, temperature change, roadwork, or cable movement. & Drift monitoring, held-out seasons and maintenance states, recalibration trigger, rollback procedure. \\
False satellite return & A low-noise but biased fix creates a global jump at exit. & Multi-epoch innovation and map checks, injected bias/spoofing tests, bounded handover transient. \\
Timing or communication fault & Offline coordinates remain accurate even though the state arrives too late for control. & Clock-offset injection, age-of-information, deadline misses, packet loss, authentication and replay tests. \\
Overconfident learned factor & Mean error is acceptable while tail risk exceeds the declared bound. & Reliability diagrams, empirical coverage, false-safe rate, conditional tail metrics, out-of-distribution tests. \\
\bottomrule
\end{tabularx}
\end{table*}

Track identity is part of position in a multi-vehicle service. Reporting only geometric matching between estimated and true trajectory sets can conceal an identity swap. Uncertainty should likewise be reported conditionally by lane, speed, vehicle class, traffic density, acoustic quality, camera visibility, tunnel segment, and maintenance state.

\section{A Validation Ladder That Constrains the Claims}
A Point of View should make future claims falsifiable. Table~\ref{tab:ladder} defines four stages. Existing roadside camera--DAS work supports portions of Stage~0, including synchronization, interpretation, and fiber-registration feasibility \cite{cohen2025video,cohen2026fibergeo}; it does not by itself validate GNSS-denied tunnel localization. Claims should advance only with corresponding evidence.

\begin{table*}[!t]
\caption{Minimum Evidence Before Progressively Stronger Claims}
\label{tab:ladder}
\centering
\footnotesize
\begin{tabularx}{\textwidth}{@{}p{0.11\textwidth}p{0.27\textwidth}p{0.29\textwidth}Y@{}}
\toprule
\textbf{Stage} & \textbf{Data and reference} & \textbf{Mandatory comparisons and stress tests} & \textbf{Supportable claim} \\
\midrule
0: Synchronized replay & Published or newly collected roadside camera/fiber/vehicle streams; verified clocks; known or independently checked fiber anchors; masked satellite intervals only for controlled analysis & Last-fix extrapolation, onboard dead reckoning, fiber only, camera only, nonlearned fusion, direct-survey versus traffic-assisted fiber registration; timestamp, anchor, and mapping perturbations & Feasibility of synchronization, fiber-channel registration, basic association, and edge replay; not a tunnel-performance or operational integrity claim \\
1: Controlled tunnel & Synchronized cameras and fiber, instrumented vehicles, surveyed geometry, independent reference trajectory & Same baselines plus oracle association, perfect fiber map, modality ablations, guarded versus immediate reacquisition & Tunnel-specific localization, lane and identity performance, calibrated uncertainty, and handover behavior for a declared controlled domain \\
2: Operational trial & Mixed traffic, multiple lanes or directions, variable vehicles and speeds, adverse visibility, repeated days or tunnels & Cross-day and cross-site generalization; sensor outage; masking; coupling change; biased satellite return; latency & Robustness and continuity within a declared operational design domain \\
3: Service readiness & Long-duration monitoring, maintenance records, communication faults, cybersecurity and privacy exercises & Availability and fallback alternatives; age of information; control-coupled safety analysis; lifecycle and rollback tests & Service availability, integrity risk, maintainability, and deployment evidence \\
\bottomrule
\end{tabularx}
\end{table*}

\subsection{Metrics}
Metrics should cover high-percentile localization and handover error, identity switches and ambiguity duration, empirical bound coverage and false-safe probability, latency and age of information, service-unavailable duration, availability, and calibration drift. Results should be conditioned on the declared operating domain rather than averaged across favorable and adverse regimes.

\subsection{Ablations and counterfactuals}
Comparisons should include no infrastructure, onboard dead reckoning, fiber only, cameras only, nonlearned fusion, oracle association, a perfect-fiber-map counterfactual, immediate satellite return, and integrity-gated return. Deliberate timestamp offsets, camera blackout, fiber loss, heavy-vehicle masking, stale maps, and biased returning satellite solutions should test whether failure is detected rather than hidden. A lower mean position error with an occasional silent lane jump may be less useful than a larger but correctly bounded error \cite{iso21448,iso26262}.

\section{Edge Deployment, Maintenance, Privacy, and Security}
Edge processing can convert high-rate fiber data and camera observations into compact state and integrity messages, reducing bandwidth while imposing timing, latency, availability, authentication, and cybersecurity requirements. Maintenance is part of the measurement model: roadwork, water ingress, fiber rerouting, camera movement, clock drift, and software changes can invalidate calibration. Health checks, versioned maps/models, recalibration, rollback, local video processing, ephemeral identities, secure timing, communications, source-health monitoring, and adversarial testing are therefore part of deployment. Where vehicle or appearance data constitute personal data, deployment must also satisfy the applicable jurisdictional requirements for lawful purpose, data minimization, retention, access, and governance; these constraints belong to the operational design domain rather than being treated as post-deployment policy.

\section{Priority Research Questions}
The position can be tested through a focused agenda: How far can anonymous fiber tracks remain attributable before another semantic anchor is needed? When is lane observable from fiber, and when is another sensor indispensable? Which calibration or adaptation methods transfer across cables, burial, pavement, geology, temperature, and interrogators? When should ambiguity remain set-valued? What statistically defensible protection level can be tied to lane geometry and downstream control? How much identity continuity can be achieved without retaining recognizable video? How should infrastructure and onboard estimates be arbitrated under latency and correlated failure? What independent reference is sufficient to validate the service?

These questions bound the opportunity. DAS is not a universal substitute for satellite navigation or onboard localization. It is most compelling where suitable fiber and edge access exist, tunnel motion is geometrically constrained, and position continuity justifies calibration and maintenance. Cameras become more valuable as traffic, lane structure, overtaking, opposite directions, and identity-sensitive services increase.

\section{Recommendations}
\subsection{For researchers}
Report synchronization error, fiber geometry, interrogator and gauge-length settings, camera calibration, preprocessing, reference uncertainty, train/test separation, and failure cases. Separate fiber-map calibration uncertainty from vehicle-localization error; include nonlearned baselines, oracle association, a perfect-map counterfactual, and calibrated uncertainty. Masked open-road satellite data should not be described as a tunnel experiment, and synthetic heat maps should not be presented as measured evidence.

\subsection{For tunnel and road operators}
Treat fiber routes, cameras, timing infrastructure, maps, maintenance records, and edge software as components of one positioning system. Begin with a bounded corridor and explicit service-level metrics. Existing cameras and communications may reduce deployment cost, but placement, compression, time stamping, calibration, and data-retention policies must be re-evaluated for localization use.

\subsection{For vehicle and standards communities}
Define interoperable semantics for coordinate frames, timestamps, track lifecycle, uncertainty, protection levels, integrity mode, source health, validity horizon, and failure notification. The goal is not identical onboard architectures, but conservative and auditable use of infrastructure evidence.

\section{Conclusion}
The opportunity is to construct a traceable service in which each modality contributes what it observes best: a trusted satellite solution anchors globally, DAS supplies continuous corridor evidence, cameras supply identity and lane anchors, maps and onboard motion constrain geometry and dynamics, and edge intelligence fuses them under an explicit integrity state.

This framing changes the research question from ``How far can a model extrapolate after satellite loss?'' to ``How can an intelligent transportation system preserve a particular vehicle's globally referenced, integrity-bounded track through a satellite blackout?'' The enabling sensors and algorithms exist, but an operational claim still requires synchronized tunnel data, independent reference truth, multi-vehicle tests, calibrated uncertainty, deliberate failure injection, latency measurement, maintenance evidence, and guarded reacquisition. Although the reference architecture is tunnel-specific, the underlying principle---attributable, integrity-bounded continuity rather than coordinate extrapolation alone---may extend to other constrained GNSS-degraded corridors; each environment would require its own sensing and validation evidence. This Point of View therefore defines a concrete research and validation agenda for moving from isolated localization components toward deployable, integrity-aware positioning continuity in intelligent transportation systems.

\section*{Data and Figure Statement}
No new experimental or synthetic performance data are reported in this Point of View. All three figures are original conceptual schematics created for this manuscript and are not presented as measured or simulated results.

\bibliographystyle{IEEEtranS}
\bibliography{references}

\end{document}